\documentclass[letterpaper, 10 pt, conference]{ieeeconf}  

\IEEEoverridecommandlockouts                              

\usepackage{cite}
\usepackage{amsmath,amssymb,amsfonts}
\usepackage{mathtools,nccmath}
\usepackage{algorithmic}
\usepackage{graphicx}
\usepackage{textcomp}
\usepackage{xcolor}
\usepackage{comment}
\usepackage{caption}
\usepackage{subcaption}
\usepackage[implicit=false]{hyperref}

\begin{document}


\title{\LARGE \bf Mechanically Programming the Cross-Sectional Shape of \\ Soft Growing Robotic Structures for Patient Transfer}

\author{O. Godson Osele$^{1}$\textsuperscript{\textdagger}, Kentaro Barhydt$^{2}$\textsuperscript{\textdagger},  Teagan Sullivan$^{2}$, H. Harry Asada$^{2}$, and Allison M. Okamura$^{1}$ 
\thanks{\textsuperscript{\textdagger}These authors contributed equally to this work and are co-first authors.}
\thanks{This work is supported in part by the Ford Foundation Predoctoral Fellowship and National Science Foundation grant $\#2024247$.}
\thanks{OGO and AMO are with the Department of Mechanical Engineering, Stanford University, Stanford, CA 94305 USA
{\tt\small \{obum\}@alumni.stanford.edu}, 
{\tt\small \{aokamura\}@stanford.edu}}
\thanks{KB, TS, and HHA are with the Department of Mechanical Engineering, MIT, Cambridge, MA, 02139 USA
        {\tt\small \{kbarhydt, cteagans, asada\}@mit.edu}}%
}

\maketitle
\thispagestyle{empty}
\pagestyle{empty}

\begin{abstract}

Pneumatic soft everting robotic structures have the potential to facilitate human transfer tasks due to their ability to grow underneath humans without sliding friction and their utility as a flexible sling when deflated. 
Tubular structures naturally yield circular cross-sections when inflated, whereas a robotic sling must be both thin enough to grow between a human and their resting surface and wide enough to cradle the human. 
Recent works have achieved flattened cross-sections by including rigid components into the structure, but this reduces conformability to the human. 
We present a method of mechanically programming the cross-section of soft everting robotic structures using flexible strips that constrain radial expansion between points along the outer membrane. 
Our method enables simultaneously wide and thin inflated profiles, and maintains the full multi-axis flexibility of traditional slings when deflated. 
We develop and validate a model relating geometric design specifications to fabrication parameters, and experimentally characterize their effects on growth rate. 
Finally, we prototype a soft growing robotic sling system and demonstrate its use for assisting a single caregiver in bed-to-chair patient transfer.

\end{abstract}

\section{INTRODUCTION}

The field of soft robotics continues to realize the promise of leveraging mechanically compliant robotic designs as technological solutions to emerging societal challenges \cite{yasa_overview_2023}. 
One potential application of such work lies in securely harnessing and transferring humans. 
In eldercare and the care for people with physical disabilities, transferring humans is a critical task regularly carried out by caregivers \cite{dewit_fundamental_2013}. 
This removes the patient's independence in daily life and can be strenuous and cause injuries for caregivers \cite{chang_impact_2010, noauthor_employer-reported_2023, feng_global_2019}. 
The current standard practice for caregivers is to manually place the patient in a sling or harness to securely lift them \cite{dewit_fundamental_2013}. 
Slings and straps are practical for high-force human interaction because they can be wrapped around the body to bear the heavy load while distributing it over a large contact area \cite{barhydt_high-strength_2023}. 
Slings are particularly beneficial, in that their wide sheet-like topology enables maximal contact area with the body. 
In current practice, caregivers must manually place the patient into and out of the sling, inflicting harmful strain on their body \cite{chang_impact_2010}. 
Such manual patient handling operations are fatiguing, dangerous, and require multiple caregivers. 

\begin{figure}[]
    \centering
    \includegraphics[trim={0.0cm 1.7cm 0.0cm 1.7cm}, clip, width=1.0\linewidth]{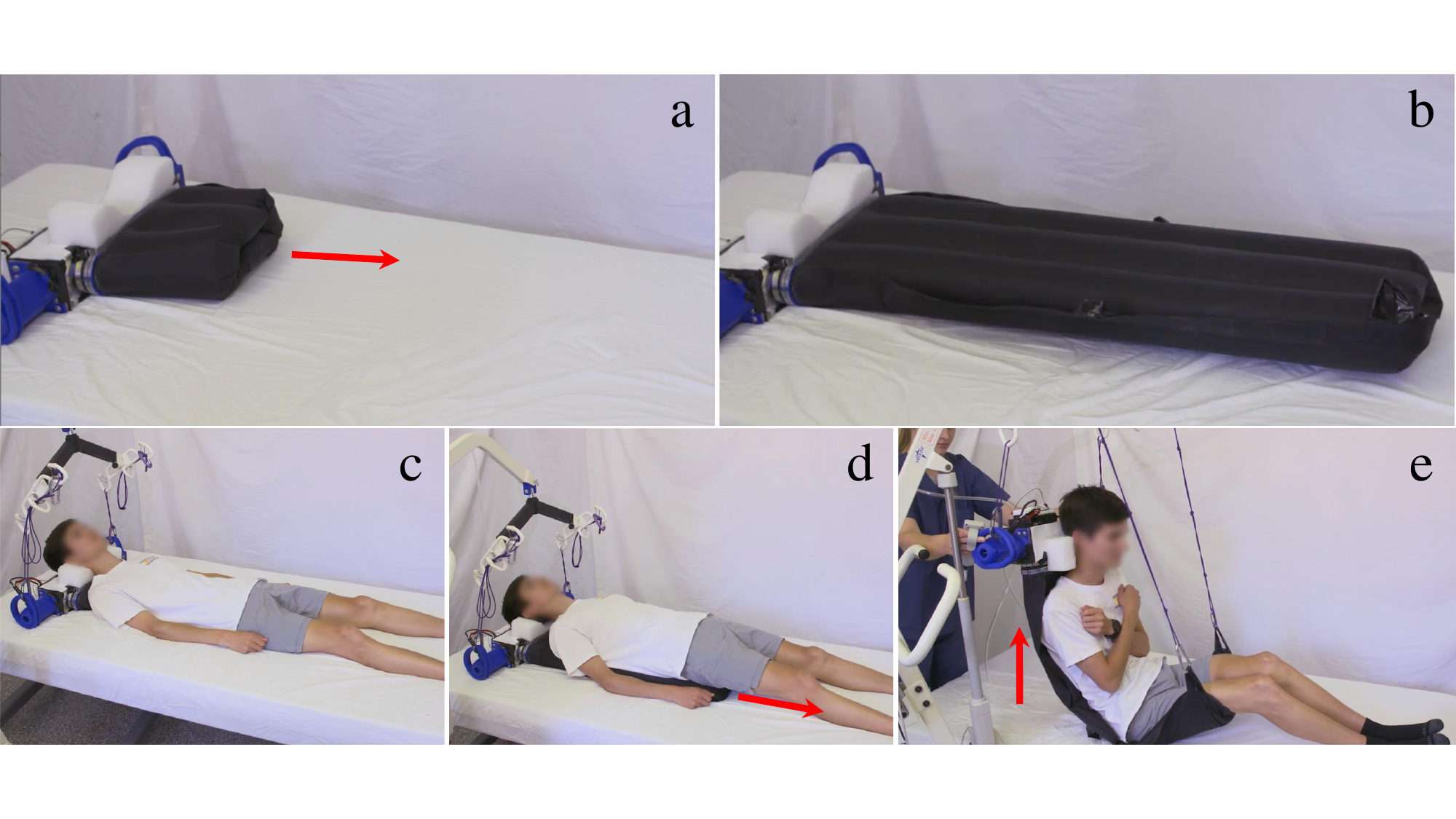}
    \caption{\fontsize{9}{9} \selectfont Soft growing robotic sling prototype with a flattened inflated cross-section. (a) and (b) show the sling growing out of a wide base and grown to its full length, respectively. (c) shows the wide base with the sling fully retracted and placed under the head of a human, (d) shows the sling fully grown and deflated under the human with its ends attached to a Hoyer lift device via cables, and (e) shows the human lifted by the sling connected to a Hoyer lift.}
    \label{fig:splashFig}
    \vspace{-5mm}
\end{figure}

While previous robots have been developed for patient transfer tasks \cite{mukai_whole-body_2011,noauthor_tmc_nodate,bolli_enhancing_2024}, traditional robotic manipulators cannot provide the simultaneously strong yet gentle interaction necessary to safely lift the full weight of the human body. 
Inspired by straps and slings used in eldercare, a semi-soft hyper-redundant robot with the high tensile strength and bending flexibility needed to gently bear high loads was previously developed, and demonstrated lifting the full weight of humans without external physical assistance \cite{barhydt_high-strength_2023}. 
However, robotic manipulators with any rigid components cannot easily be placed between the patient and their resting surface without an external party adjusting and/or lifting their body, primarily due to the lack of a gap to enter and the resulting sliding friction. 
Recent literature has demonstrated soft robotic designs leveraging eversion to grow (and retract from) under the patient without sliding friction \cite{nakamura_soft_2018,choi_development_2020
}. 
\cite{nakamura_soft_2018} utilizes pneumatically driven eversion to insert soft cylindrical manipulators underneath the human body to aid with turning. 
This work is limited to a partial assist as opposed to a full body patient transfer. 
Towards realizing the employing of soft everting structures for full body transfer, \cite{choi_development_2020} presented a pneumatically driven growing sling that restricted the radial expansion of an inflated beam to achieve a flatter shape under the patient, better suited for their comfort, by integrating rigid shafts into the membrane. 
These rigid shafts restrict the sling's flexibility to bending only about the patient's frontal axis, and limit the cross-sectional profile to convex shapes. 
The authors of this paper have previously demonstrated lifting the full weight of a human with multiple fully soft growing robots without any rigid components \cite{barhydt2025loopclosuregraspingtopological}. 
However, the growing robots had a cylindrical cross-section and were used as individuals straps working together to support the human along their upper back and under their knees, unlike traditional patient slings which can be wide enough to better securely support the entire body with just a single sheet. 

In this paper, we present a method for mechanically programming the cross-sectional shape of soft, everting, compliant robotic structures and realize a self-tunneling and retracting sling. 
We present the first demonstration of safely harnessing and transferring the full weight of a human with a robotic sheet-shaped sling device used by a single operator. 
Our robotic sling extends and retracts under the body automatically with no sliding friction, eliminating the uncomfortable experience of being rolled back and forth. 
Once fully deployed and deflated, the design maintains high global all-axis bending compliance about all body axes (frontal, sagittal, longitudinal) with high tensile integrity by employing a simple fabrication process that includes only heat-sealable fabric. 
We also present and validate a model to select the fabrication parameters for a desired cross-section design. 
Fabrication methods are developed to realize our design approach, and a prototype of a soft growing pneumatic sling with full multi-axis bending flexibility is designed and built for patient transfers. 
The system is experimentally characterized, demonstrating its ability to harness and lift humans effectively for bed-to-chair transfers. 

\section{DESIGN}

\subsection{Principles}


\subsubsection{Eversion}
Our system achieves “growth” and “retraction” at its tip through pressure-driven eversion of flexible, thin-walled membranes. 
This process, first presented in \cite{hawkes_soft_2017}, refers to starting with the flexible, thin-walled, inextensible membranes inverted and achieving lengthening from the tip by pressuring the system to transfer material from within its tubular body to the outside. 
The growth rate of the robotic structure is additionally constrained by a ``tail'' (e.g. a string) attached to the end of its everted material. 
Control of the internal pressure and string release is required for precise control of the system’s growth.
This eversion principle coupled with the compliance of the membranes enables soft growing robots to achieve high extension ratios and navigate through cluttered environments without external surface friction. 
Specific to our application, it allows for navigation through gaps that are significantly smaller than its width, which is required to insert a sling underneath a patient in a supine position. 
By using eversion, the flexible membrane can grow into small gaps between the subject’s body and the bed surface without inducing damaging sliding friction to the skin and wounding the patient. 
The lifting force needed for soft growing robotic structures leveraging eversion to grow into the slanted gap underneath the patient has been studied in past literature \cite{hawkes_soft_2017, blumenschein_modeling_2017, choi_development_2020}, and shown to be primarily dependent on the internal air pressure and the angle of the slanted gap.

\subsubsection{Restricting radial expansion}

\begin{figure}[]
    \centering
    \includegraphics[trim={0.0cm 1.2cm 0.0cm 0.0cm}, clip, width=1.0\linewidth]{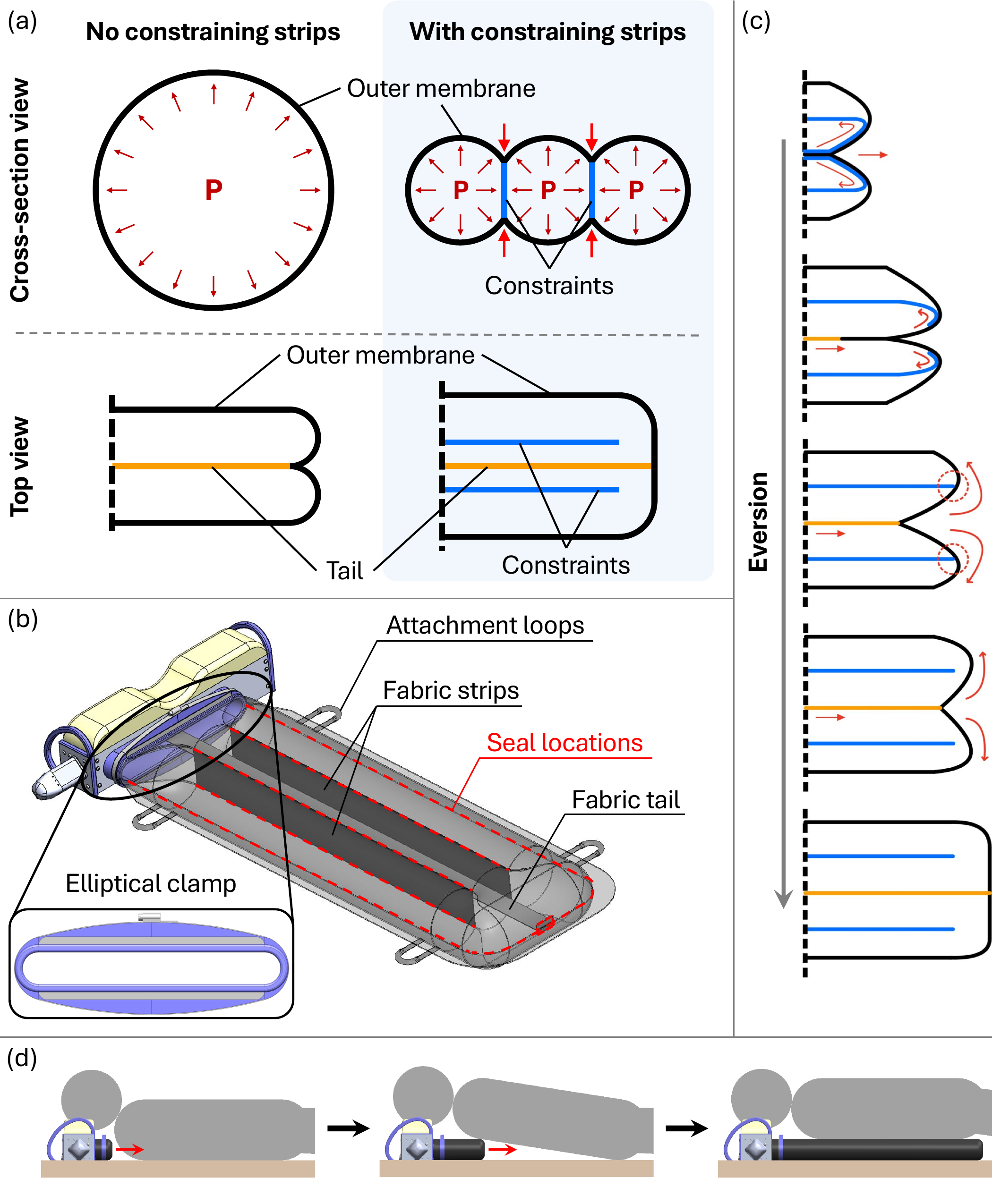}
    \caption{\fontsize{9}{9} \selectfont (a) Constraining strips pulls points along the outer membrane inward to flatten the cross-section. (b) Full design of the soft growing robotic sling system, including the robotic sling with flattened cross-section, a wide motorized base to accommodate the flattened cross-section, and loops to attach to Hoyer lift cables. (c) Sequence over which the robotic sling with constraining fabric strips everts. (d) Sequence of robotic sling growing between the human body and its resting surface.}
    \label{fig:prototype}
    \vspace{-5mm}
\end{figure}

The ideal geometry for a structure meant to grow underneath patients is one that is wide enough to span the patient’s body and flat enough to avoid discomfort. 
The inflated geometry of soft everting growing robots, barring external forces, is governed by its wall tension when pressurized. 
By systematically fabricating the robot body such that it experiences wall tension that restricts its radial expansion into a desired shape, we can design the everting body to have a flatter inflated geometry. 
The first soft growing robotic structure flat enough to be applicable as a growing sling for patients was described in \cite{choi_development_2020}. 
This design restricts radial expansion by integrating shafts into the flexible membrane serving as the robot body. 
The shafts provide an antagonistic moment for the radially expansive forces and are aligned such that they do not interfere with eversion. 
However, this design results in a loss of all-axis bending flexibility that allows for better conformation of the sling to the body. 
In our design, we address this by using only flexible fabric. 
As shown in Fig. \ref{fig:prototype}(a), we bond pairs of points along the robot body’s cross-sectional circumference perpendicular to the direction of its net eversion growth to constrain its radial expansion when inflated. This is akin to how some air mattress designs achieve their flat surface topology. 
To make this design viable for the eversion principle, the bonded points must allow for a middle channel for the flexible membrane to evert from and sufficiently low capstan friction from the internal materials sliding against each other \cite{blumenschein_modeling_2017, haggerty_characterizing_2019}.  

\subsection{Implementation and Fabrication}

\subsubsection{Fabricating the robot body}
To implement restriction of radial expansion, we used thermoplastic-coated fabric (TPU-coated ripstop nylon) due to its exceptional tensile strength and ease of forming robust bonds through heat-sealing the coated sides using an ultrasonic welder (VETRON 5064). 
The process involved cutting the fabric material that forms the robot body’s perimeter. 
Subsequently, two strips of fabric were cut to span the net growth length of the robot body, as depicted in Fig. \ref{fig:prototype}(b). 
Each strip was heat-sealed along two predetermined lines on the perimeter fabric material in the direction of the robot body’s net eversion growth. 
The distance between these lines was prescribed to constrain the radial expansion by a predetermined amount. 
The calculation of this distance is detailed in Section \ref{modeling}. 
Finally, the perimeter fabric material was folded to realize the form shown in Fig. \ref{fig:prototype}(b), ensuring that the TPU-coated side contacts itself so it can be heat-sealed along its edges to create an airtight enclosure for the robot body.

\subsubsection{Creating strong attachments}
The size and load capacity of the robotic sling structure must be great enough to harness and lift the human body, leading to points on the membrane that experience high force concentrations. 
First, the tail applies high forces onto the inner tip of the membrane when inflated, as the large cross-sectional area leads to high pressure-driven eversion forces (final prototype has cross-sectional area of $3.67\times10^4$ mm$^2$ and driven at 34 kPa, yielding 1249 N eversion force). 
Thus, our prototype utilizes a wide strip of fabric (same as membrane) as the tail instead of a standard string. 
The end of the strip is welded to the inner tip of the structure, such that the holding force between them is distributed across a wide seam instead of concentrated at a single point. 

The fastening points between the Hoyer lift cables and the membrane also creates force concentrations when lifting the human body. 
While most commercial patient slings have webbing loops sewn onto their edges for the cables to attach to, sewing onto the membrane introduces punctures that lead to air leaks and/or tears. 
Thus, our prototype implements tubular loops of fabric (same as membrane) welded to the membrane, as shown in Fig. \ref{fig:prototype}(b). 
The loops are welded inside of the membrane's outer seam such that the welds are pulled in shear instead of peeling for increased strength. Fabricated as such, each of the loops can bear over 68~kg.


\subsubsection{Wide base}
The base and collar were designed with a long, narrow shape to match the flattened cross-section of the robotic sling, as well as enable the base to be placed under the human’s head. 
As shown in Fig. \ref{fig:prototype}(b), a hose clamp fastens a set of curved blocks around the collar, which clamp the robotic sling membrane to the collar. 
Because the normal pressure the hose clamp applies at a given point is proportional to its local curvature, the blocks are designed such that the hose clamp holds a roughly elliptical profile (as opposed to the slot-shaped collar) and applies some normal pressure around the entire perimeter without excessively protruding from the long and narrow shape of the base.

\section{MODELING}
\label{modeling}

We present an analytical model to determine the fabrication parameters required to meet design specifications. 

\begin{figure}[]
    \centering
    \includegraphics[trim={0.0cm 4.2cm 0.0cm 0.0cm}, clip, width=0.9\linewidth]{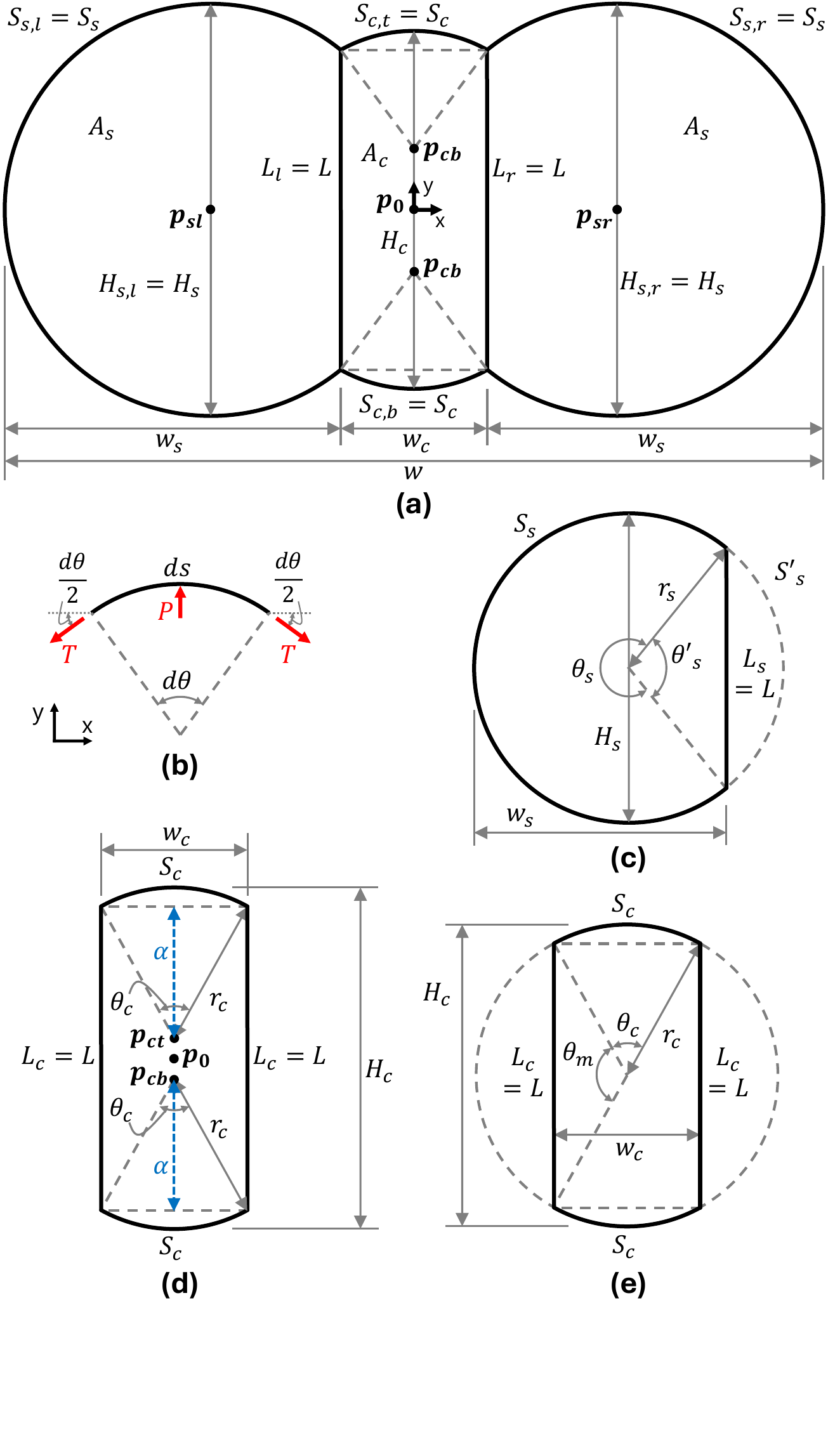}
    \caption{\fontsize{9}{9} \selectfont Geometry of robotic sling cross section. (a) Full cross section geometry. (b) Free-body diagram of differential segment of curved section. (c) Side channel geometry. (d) Center channel geometry without aligned arc centers. (e) Center channel geometry with aligned arc centers.}
    \label{fig:modeling}
    \vspace{-5mm}
\end{figure}

\subsection{Problem Definition and Assumptions}

The robotic sling cross-section geometry is shown in Fig. \ref{fig:modeling}(a). 
We define the design specifications as the heights of each channel and the overall width of the structure when inflated ($H_{s,l}$, $H_{s,r}$, $H_c$, and $w$; subscripts $r$ and $l$ denote left and right channels, respectively). 
Higher side channels help keep the body on the robotic sling stable lying on the center channel, but decreasing the channel size reduces the space through which the material can evert, and thus a design must balance these tradeoffs to enable stable placing while enabling eversion and inversion. 
Additionally, the robotic sling width must be designed according to the width of the target bodies.   
We define the fabrication parameters as the lengths of each segment of the cross-section ($S_{c,t}$, $S_{c,b}$, $S_{s,l}$, $S_{s,r}$, $L_r$, and $L_l$; subscripts $t$ and $b$ denote top and bottom segments, respectively). 
Unlike the design specifications, these fabrication parameters can be directly controlled during the fabrication process by cutting and welding the membrane to have desired lengths. 

In this work, we assume that the cross-section of the robotic sling is uniform across its length, the cross-section geometry is symmetrical across its central vertical and horizontal axes (x and y axes in Fig. \ref{fig:modeling}(a)), and the membrane has negligible flexural rigidity and is inextensible. 
Given the symmetry, the set of design specifications can be reduced to $H_s=H_{s,l}=H_{s,r}$, $H_c$, and $w$, and the set of fabrication parameters can be reduced to $S_c=S_{c,t}=S_{c,b}$, $S_s=S_{s,l}=S_{s,r}$, $L=L_l=L_r$. 
Additionally, the symmetry implies that the straight segments must be parallel to each other, their midpoints lie on the x-axis, the centers of the top and bottom arcs lie on the y-axis, and the centers of the side arcs lie on the x-axis.

\subsection{Simplifying Geometric Principles}

Here, we present two principles that we apply to simplify the parameterization and analysis of the inflated cross-section geometry: (1) all arc segments have a constant curvature, and (2) the top and bottom arcs of the center channel are coradial. 

\subsubsection{Curved segment parametrization}

Given that the membrane is inextensible and has negligible flexural rigidity, and its mechanics can be analyzed in the cross-sectional plane due to the uniform cross-section of the robotic sling, the curved segments of the cross-section geometry are assumed to have a constant curvature.  

A free-body diagram of a differential segment of the membrane is shown in Fig. \ref{fig:modeling}(b). 
The differential force applied onto the segment from the internal pressure (aligned with the y-axis) is balanced by the tension in the membrane. 
Since the membrane is inextensible and has negligible flexural rigidity, the only forces that the segment experiences are tension and air pressure, both of which are constant across the membrane length. 
The force balance is as follows:
\begin{equation}
    \sum F_x=0=T\cos{\frac{d\theta}{2}}-T\cos{\frac{d\theta}{2}}
    \label{eq:diffSegmentFBD_x}
\end{equation}
\begin{equation}
    \sum F_y=0=Pds-2T\sin{\frac{d\theta}{2}}
    \label{eq:diffSegmentFBD_y}
\end{equation}
where $T$ is the tension, $P$ is the air pressure, $ds$ is the differential arc length, and $d\theta$ is the central angle of $ds$. 
Given that $d\theta$ is small, Eq. \ref{eq:diffSegmentFBD_y} is reduced to:
\begin{equation}
    \frac{P}{T}=\frac{d\theta}{ds}=\kappa
    \label{eq:diffSegmentFBD_curvature}
\end{equation}
where $\kappa=\frac{d\theta}{ds}$ is the curvature of the segment. 
Both $P$ and $T$ are constant across the segment length, thus $\kappa$ is constant. 
Therefore, all curved segments have a constant curvature, and can be modeled as circular arcs.

\subsubsection{Arcs of center channel are coradial}

Here we show that the top and bottom arcs of the center channel share the same center at the origin of the cross-section. 
Fig. \ref{fig:modeling}(d) shows a diagram of the cross-section center channel geometry with different arc centers. 
Given only the design specifications and fabrication parameters, this geometry is indeterminate, as the central angle $\theta_c$ can have any value. 
By showing that the centers of the arcs must be aligned, we introduce an additional constraint that enables the geometry to be fully determined given only the design specifications and fabrication parameters. 

The centers of the top and bottom arcs and the origin are denoted as $\mathbf{p_0}=\left[p_{0,x}, p_{0,y}\right]^T$, $\mathbf{p_{ct}}=\left[p_{{ct},x}, p_{{ct},y}\right]^T$, and $\mathbf{p_{cb}}=\left[p_{{cb},x}, p_{{cb},y}\right]^T$.
Given symmetry across the x- and y-axes, $p_{0,x}=p_{ct,x}=p_{cb,x}=0$.
We also define the term $\alpha=r_c\cos{\frac{\theta_c}{2}}$.
If $2\alpha=L$, then $p_{0,y}=p_{ct,y}=p_{cb,y}$, and thus $\mathbf{p_{ct}}$ and $\mathbf{p_{cb}}$ are aligned with $\mathbf{p_{0}}$.

A governing principle for the geometry of inflated structures is that, assuming inextensibility and no flexural rigidity, the membrane is deformed such that the internal volume is maximized ($A_c=A_{c,max}$). 
Thus, $\frac{dA_c}{d\theta_c}=0$ must be true for any $S_c$, $L$, and $\theta_c$ (geometric parameters which fully determine the cross-section geometry). 
If $2\alpha=L$ must be true for $A_c=A_{c,max}$ to be true for any $S_c$, $L$, and $\theta_c$, then $p_{0,y}=p_{ct,y}=p_{cb,y}=0$.
The cross-sectional area and its derivative with respect to $\theta_c$ are expressed as:
\begin{equation}
    A_c=\frac{S_c^2}{\theta_c}-\frac{S_c^2}{\theta_c^2}\sin{\theta_c}+\frac{2S_c}{\theta_c}\sin{\frac{\theta_c}{2}}L
    \label{eq:A_c}
\end{equation}
and, 
\begin{align}
    \frac{dA_c}{d\theta_c} = \frac{2S_c}{\theta_c^2} \left(2\sin{\frac{\theta_c}{2}}-\theta_c\cos{\frac{\theta_c}{2}}\right) \left(\frac{S_c}{\theta_c}\cos{\frac{\theta_c}{2}}-\frac{L}{2}\right).
    \label{eq:dA_c/dtheta_c 2}
\end{align}
The first and second terms of Eq. \ref{eq:dA_c/dtheta_c 2} (denoted $f_1\left(\theta_c\right) = \frac{2S_c}{\theta_c^2}$ and $f_2\left(\theta_c\right) = 2\sin{\frac{\theta_c}{2}}-\theta_c\cos{\frac{\theta_c}{2}}$, respectively) are non-zero when the area is maximized. 
$f_1\left(\theta_c\right)=0$ if and only if $S_c=0$.
However, if $S_c=0$, then $A_c=0$ according to Eq. \ref{eq:A_c}, and thus $A_c\neq A_{c,max}$.
For $f_2\left(\theta_c\right)$, $f_2\left(\theta_c\right)>0$ for $0<\theta_c<2\pi$. 
The admissible range of $\theta_c$ for $A_c=A_{c,max}$ to be true is $0<\theta_c<2\pi$ ($\theta_c\neq0$ because $\lim_{\theta_c \to 0}{A_c}=0$).
Thus, for $\frac{dA_c}{d\theta_c}=0$ and $A_c=A_{c,max}$ to be true for any $S_c$, $L$, and $\theta_c$, $f_3\left(\theta_c\right)=S_c\theta_c^{-1}\cos{\frac{\theta_c}{2}}-\frac{1}{2}L=0$ must be true for any $S_c$, $L$, and $\theta_c$.
Given $r_c=S_c\theta_c^{-1}$, $f_3\left(\theta_c\right)=0$ is equivalent to:
\begin{align}
    2r_c\cos{\frac{\theta_c}{2}}=2\alpha=L
    \label{eq:2alpha}
\end{align}
for $0<\theta_c<2\pi$.
Therefore, $2\alpha=L$ must be true for $\frac{dA_c}{d\theta_c}=0$ to be true for any $S_c$, $L$, and $\theta_c$.
Thus, $p_{0,y}=p_{ct,y}=p_{cb,y}=0$ and $\mathbf{p_0}=\mathbf{p_{ct}}=\mathbf{p_{cb}}=\left[0, 0\right]^T$, i.e. the top and bottom arcs are concentric. 
Because of this concentricity and the fact that these arcs have equal radii due to symmetry, they are coradial as well.

\subsection{Cross-Section Geometry Model} \label{sec:geometryModel}

With these principles, we derive a fully analytical model to express the fabrication parameters in terms of the design specifications. 
For brevity, we present only the results of the derivation here, and the full derivation is presented in the appendix. 
In short, we analyze and define the geometry of the side and center channels in terms of the variables shown in Fig. \ref{fig:modeling}(c) and Fig. \ref{fig:modeling}(e), respectively. 
We then analyze the entire cross-section geometry as a whole (Fig. \ref{fig:modeling}(a)) by equating the side and center channel geometries using $w = w_c + 2 w_s$ and $L = L_c = L_s$, which then enables the derivation of equations for the fabrication parameters $L$, $S_c$, and $S_s$, in terms of only the design specifications $H_c$, $H_s$, and $w$. 
The resulting equations are 
\begin{align}
    S_c=H_c\sin^{-1}{\left(\frac{w^2+H_c^2-2wH_s}{2H_c\left(w-H_s\right)}\right)},
    \label{eq:S_c}
\end{align}
\begin{align}
    L=H_c\sqrt{\frac{\gamma}{4H_c^2\left(w-H_s\right)^2}}
    \label{eq:L}
\end{align}
where 
\begin{align}
    \gamma = 4H_s (H_c^2 H_s - H_c^2 w - w^2 H_s + w^3) - (w^2 - H_c^2)^2 ,
    \label{eq:gamma}
\end{align}
and 
\begin{equation}
    S_s = 
    \begin{cases} 
        \beta & \text{if } w - H_c\sin{\frac{S_c}{H_c}} \le H_s \\ 
        \pi H_s - \beta & \text{if } w - H_c\sin{\frac{S_c}{H_c}} > H_s
    \end{cases}
    \label{eq:S_s}
\end{equation}
where 
\begin{align}
    \beta = H_s \sin^{-1} \Biggl( \frac{H_c}{H_s} \sqrt{ \frac{ \gamma }{4H_c^2\,(w - H_s)^2} } \Biggr).
    \label{eq:beta}
\end{align}

Additionally, given the fabrication parameters, the full geometry of the cross section can then be determined.
The radii of the curved segments of the side and center channels are given as $r_s = H_s/2$ and $r_c = H_c/2$, respectively.
The centers of the side arcs are $\mathbf{p_{sr}}=\left[p_{sr,x}, p_{sr,y}\right]^T$ and $\mathbf{p_{sl}}=\left[p_{sl,x}, p_{sl,y}\right]^T$, where $p_{sr,x}=\frac{w}{2}-r_s$, $p_{sl,x}=-\frac{w}{2}+r_s$, and $p_{sr,y}=p_{sl,y}=0$.
The centers of the top and bottom arcs are already defined previously as $\mathbf{p_{ct}}=\mathbf{p_{cb}}=\mathbf{p_0}=\left[0, 0\right]^T$.
The midpoints of the straight segments are given as: $\mathbf{p_{Lr}}=\left[\frac{w_c}{2}, 0\right]^T$ and $\mathbf{p_{Ll}}=\left[-\frac{w_c}{2}, 0\right]^T$, where the width of the center channel $w_c$ is given as 
\begin{align}
    w_c=2r_c\sin{\frac{\theta_c}{2}}=H_c\sin{\frac{S_c}{H_c}}.
    \label{eq:w_c}
\end{align}

Matlab scripts for the analytical model, along with a numerical model that directly yields the cross-sectional geometry from a given set of fabrication parameters, is provided in the following Github repository: \url{https://github.com/kentaro-barhydt/softGrowingRobotCrossSectionProgramming}.

\section{EXPERIMENTS AND DEMONSTRATIONS}



\begin{figure}[]
    \centering
    \includegraphics[trim={0.0cm 0.0cm 0.0cm 0.0cm}, clip, width=1.0\linewidth]{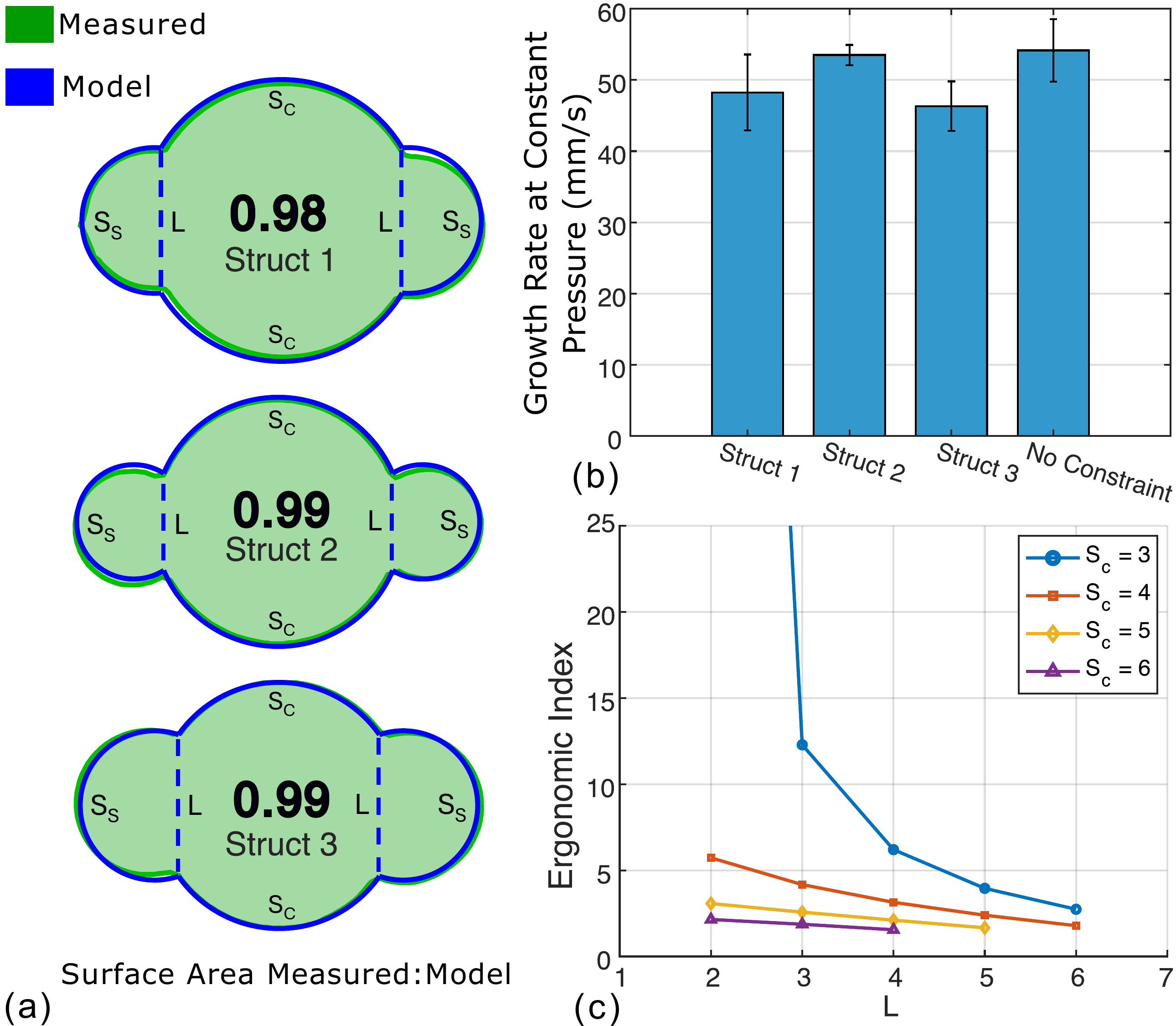}
    \caption{\fontsize{9}{9} \selectfont 
    (a) Geometry model cross-section shape prediction vs. experimentally measured prototype shape. Measurement to model area ratios ($A_{measured}:A_{model}$) ranged from 0.98 to 0.99. 
    (b) Growth rates of prototypes with different fabrication parameter values driven at constant pressure.
    (c) The effects of fabrication parameters on user comfort.}
    \label{fig:experiments}
    \vspace{-5mm}
\end{figure}

\subsection{Geometry Model Validation}
\label{modelval}

To validate the analytical model predicting the inflated cross-sectional shape of the soft growing robotic structures, we fabricated three iterations of the robot varying the fabrication parameters. 
The iterations are as such: Structure 1 ($S_c$: 152 mm, $S_s$: 127 mm, $L$: 76.2 mm), Structure 2 ($S_c$: 152 mm, $S_c$: 127 mm, $L$: 50.8 mm), Structure 3 ($S_c$: 127 mm, $S_s$: 152 mm, $L$: 76.2 mm). 
Each structure was attached to an airtight pressurized base and fully inflated at 2.07 kPa. 
We use a closed loop pressure regulator (QB4 Electro-Pneumatic Pressure Regulator, Proportion-Air, Inc.) and wireless pressure sensor (Pasco Wireless Pressure Sensor 3203) to control and monitor the robot body internal pressure. 
While fully inflated, a 3D Scanner (Creality Scan-01) was used to capture the 3D geometry of the robot body as a mesh. 
The captured mesh data was then post-processed using a CAD software to normalize and compare the measured surface area to the model predicted surface area using the same fabrication parameters. 
The results are presented in Fig. \ref{fig:experiments}(a). 
The model is in good agreement with the measured data, as demonstrated by an average measurement to model ratio of 0.99.

\subsection{Effects of Fabrication Parameters on Growth Rate}
\label{growthrate}

Recognizing that this design approach limits the space inside the robot body for new material to evert through thus increasing the contact area between the internal everting material and external everted material and resultantly the capstan friction experienced by the system, we performed this experiment to investigate the effects of the fabrication parameters, $S_{c}$ and $L$ on the growth rate of the soft growing robotic structure at a constant pressure. 
We fabricated three iterations of the robot body using our design approach holding their perimeter distance at a constant $559$ mm but varying the fabrication parameters holding one of each fabrication parameter constant in at least two of the iterations. 
The iterations are the same as those used in the geometry model validation experiments.
Structure 2 holds the same $S_{c}$ as Structure 1 but a smaller $L$ to study the effects of $L$ on growth rate. 
Structure 3 holds the same $L$ as Structure 1 but a smaller $S_{c}$ to study the effects of $S_{c}$ on growth rate. 
To provide a basis of comparison to previously presented soft growing robots without added bonds along their cross-sectional perimeter \cite{hawkes_soft_2017}, we also fabricated and tested a tubular body structure of the same perimeter distance as the other structures with an unconstrained radial expansion. 
Each structure was $914$ mm in length and attached to an airtight pressurized base. 
We use the same pressure regulator and pressure sensor used in the geometry model validation experiments to control and monitor the robot body internal pressure. 
As a preliminary step, the robot body is everted to its full length, and then inverted by $432$ mm to establish the starting position. 
Before each trial, the robot body is inverted back to the starting position. 
For each trial, the robot growth is physically impeded while it is inflated to a starting pressure of 2.07 kPa. 
Once the starting pressure is reached, the robot growth is unobstructed and the robot is allowed to evert until it stops. 
To limit the effects of fabric wrinkling on the growth rate, we chose a pressure level sufficiently high enough for the robot body to consistently grow to complete eversion without stopping. 
We perform three trials for each robotic structure. 
Each trial was filmed with a high-definition camera at 30 frames per second, and the video was post-processed using MATLAB's Computer Vision Toolbox to track the growth rate of the robotic structures. 
The results are presented in Fig. \ref{fig:experiments}(b). 


On average, Structure 2 was $10.9\%$ faster than Structure 1, suggesting that a lower $L$ reduces impedance to the growth rate of the structure. 
In contrast, Structure 3 was $4.0\%$ slower than Structure 1, indicating that a lower $S_c$ increases impedance to the growth rate of the structure. 
Equations \ref{eq:w_c} and \ref{eq:L_c} show that there is a direct trade-off between the physical properties of $L_c$ and $w_c$, where lowering $S_c$ results in a smaller $w_c$ and a higher $L_c$. 
At $L_c^{\text{max}}$, $w_c$ is minimized to 0, leaving no channel in the middle for material to evert from. 
Alternatively, maximizing $w_c$ reduces $L_c$ to 0, but a middle channel remains for material to evert through. 
The behavior conveyed by the data suggests that it is more beneficial to the growth rate to trade off less $L_c$ for more $S_c$, and thus more $w_c$. 
This phenomenon occurs despite Equations \ref{eq:A_c}, \ref{eq:w_c}, and \ref{eq:L_c} demonstrating that a reduction in $L_c$ yields a greater reduction in $A_c$ compared to a congruent reduction in $w_c$.

As expected, the tubular body structure with an unconstrained radial expansion everts the fastest, because it maximizes the internal space allowed for new everting material to pass through. 
Compared to the programmed structures with constrained radial expansion, this allows for less surface contact and therefore friction between the internal everting material and external everted material to impede the growth rate. 
In general, we note that the growth rate of all the structures remains on the same order of magnitude, with the unconstrained structure being $14.5\%$ faster than the slowest programmed structure. 
This suggests that though there is a drop in growth rate, the design remains in a practical range for leveraging eversion technology.


\subsection{System Deployment}

\begin{figure}[]
    \centering
    \includegraphics[trim={0.0cm 0.0cm 5.7cm 0.0cm}, clip, width=1.0\linewidth]{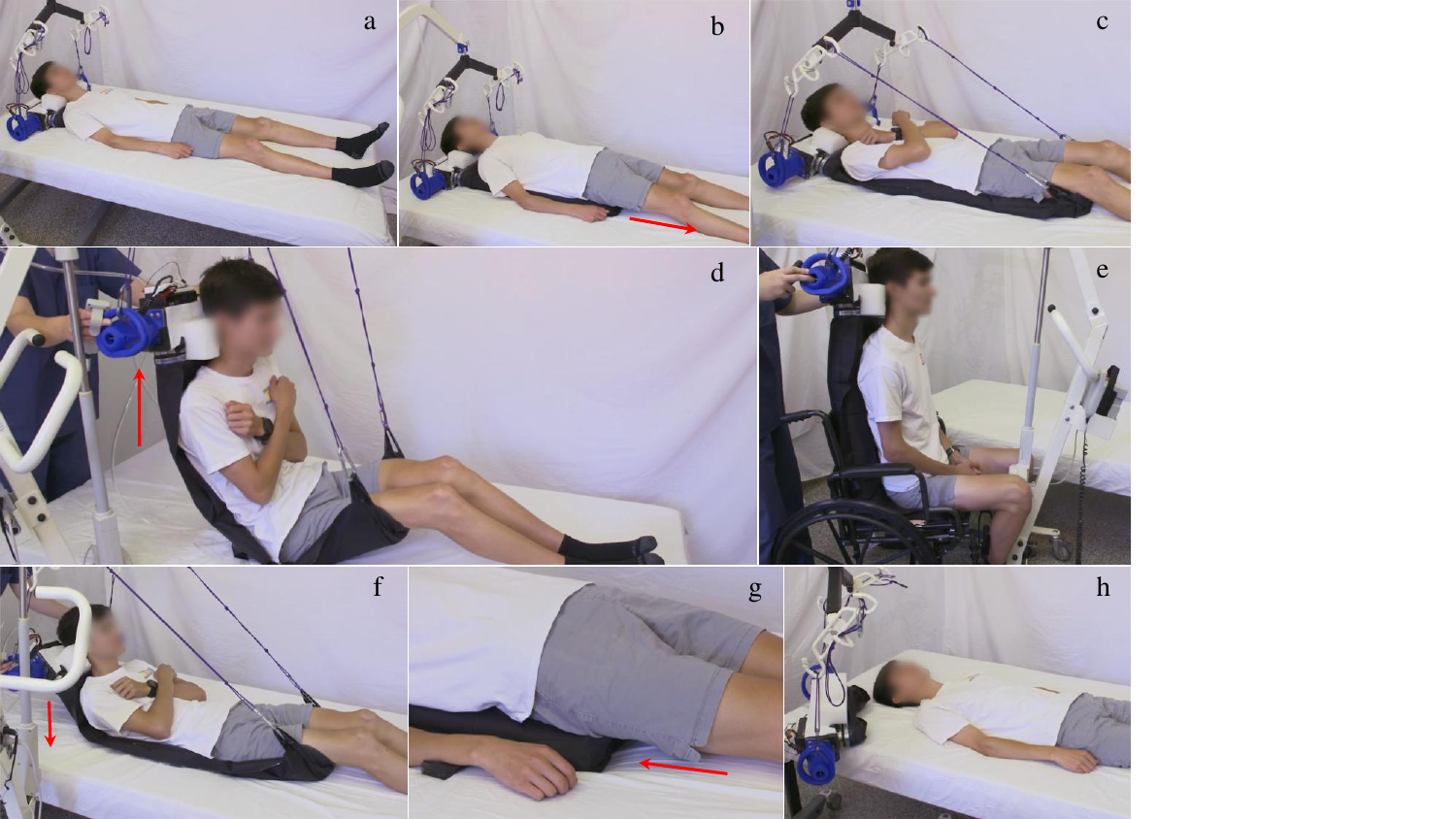}
    \caption{\fontsize{9}{9} \selectfont Patient transfer demonstrations. (a-b) With the soft growing robotic sling system placed under the participant's head, the robotic sling grows underneath the participant's body. (c-d) Participant is lifted from the bed with the deflated robotic sling using a Hoyer lift. (e) Hoyer lift is maneuvered by the caregiver to transfer the participant to a chair. (f) Participant is transferred back to lay in the bed. (g) Robotic sling automatically retracts itself from under the participant. (h) The system, including the wide base, is manually removed from under the participant's head.}
    \label{fig:demo}
    \vspace{-5mm}
\end{figure}

To demonstrate the effectiveness of our system, the prototype was used to harness and transfer human subjects from a bed to a chair, and then back to the bed. 
This demonstration was performed with five participants (5 adult males, age 19-32, weight 56-98 kg). 
As shown in Fig. \ref{fig:demo}, with the participant lying prone on a bed, a caregiver placed the base with the robotic sling fully inverted underneath their head. 
The base was inflated to 34 kPa (5 psi) and the robotic sling was everted to grow underneath the participant.
Once grown to the full length of their body, the robotic sling was then deflated to assume a sheet-like form, and the caregiver then attached the loops on the robotic sling to the cables on a Hoyer lift.
The Hoyer lift was then operated to lift the patient from the bed and move them into a chair, using standard procedures used in caregiving industry. 
To reverse the transfer, the participant was lifted from the chair and moved back to lay them onto the bed. 
The robotic sling was then inflated and inverted to remove it from underneath the participant, after which the caregiver removed the base from under the participant’s head.
The study was approved by the Massachusetts Institute of Technology’s Internal Review Board, and participants gave informed consent (participants appearing in the supplemental video gave informed consent to allow their faces to be visible). 
The prototype was successful in completing every step of the demonstration for all participants, in that it automatically placed the robotic sling under the participant’s body without human intervention, it enabled the participant to be lifted and transferred to and from a chair with the same procedure used in caregiving industry, and the robotic sling was then automatically removed from under the participant without human intervention. 


As a method of quantifying the comfort of patients while the robot is deployed underneath them, we define an ergonomic index as the inverse of the slope between the highest point of the $A_s$ channel and the highest point of the $A_c$ channel (denoted in Fig. \ref{fig:modeling}(a)). 
As intended by the design, a horizontal slot cross-sectional shape allows for a flatter surface for the patient to lay on by minimizing discomfort caused by the rounding of the patient’s back that is present with cylindrical soft growing robotic structures.
As defined above, a higher ergonomic index trends towards a concave cross-sectional profile of the robotic sling to better achieve a horizontal slot shape. 
To investigate the effects the fabrication parameters have on the ergonomic index, we use our geometric model to simulate a sweep of ergonomic indices when holding the structure’s perimeter distance constant while varying the fabrication parameters. 
The results are presented in Fig. \ref{fig:experiments}(c). 
The data suggests that, for parameter values that yield physically reasonable shapes, decreasing $L_c$ while holding $S_c$ constant increases the ergonomic index. 
It also suggests that decreasing $S_c$ while holding $L_c$ constant increases the ergonomic index. 
For each combination of  $S_c$ and $L_c$, there exists a cross-sectional perimeter value that maximizes the ergonomic index which practitioners of this design approach can leverage the analytical model to find.

\section{CONCLUSION}

This paper presents a method of mechanically programming the cross-sectional shape of soft everting robotic structures using flexible strips that constrain the radial expansion between points along the outer membrane. 
This method is the first to enable these capabilities while maintaining the full multi-axis bending flexibility of the membrane when deflated. 
We also present methods for fabricating and implementing these structures for robotic slings, develop and validate an analytical model relating its design specifications and fabrication parameters, and prototype a full soft growing robotic sling system to demonstrate its use for assisting a single caregiver in harnessing and transferring a patient from a bed to a chair. 
Unlike traditional slings, our prototype places itself under the patient automatically via eversion, eliminating the need for the caregiver to do so manually. 
Due to its multi-axis bending compliance once deflated, it could then subsequently be used exactly like a standard sling to perform the lift and transfer. 
Our demonstration is the first to harness, lift, and transfer a patient using a soft growing robot with a sling-like flattened cross-section that distributes contact pressures over a wide area.
Future work will also include the integration of bend mechanisms to enable eversion/retraction through bent pathways (e.g. under a patient sitting on a chair), and further generalization of the geometric model to account for non-symmetrical designs.
There is also room for more in-depth modeling and experimental characterization of our system under different external loads during growth and retraction, as well as further user studies with larger and more diverse participant sample sizes. 
Although the participants did not need to lift or support their upper body to aid deployment, a future study could include soft contact sensing to better understand the safety and adaptability of our prototype to scenarios with less responsive users.

\section*{APPENDIX}

Here we present the full derivation of the analytical model presented in Section \ref{sec:geometryModel}. 
We first define the geometry of the side and center channels, as illustrated in Fig. \ref{fig:modeling}(c) and Fig. \ref{fig:modeling}(e), respectively. 
The two side channels are symmetric, and thus only one must be analyzed. 
The radius and subsuming angle of the side channel arc are expressed in terms of the design specifications as 
\begin{align}
    r_s=\frac{1}{2}H_s
    \label{eq:r_s}
\end{align}
and 
\begin{align}
    \theta_s=\frac{S_s}{r_s}=\frac{2S_s}{H_s},
    \label{eq:theta_s}
\end{align}
respectively. 
We also define the conjugate arc $S^\prime_s$, as well as its subsuming angle $\theta_s^\prime=2\pi-\theta_s$ as 
\begin{align}
    \theta_s^\prime=\frac{S_s^\prime}{r_s}=\frac{2S^\prime_s}{H_s}.
    \label{eq:theta_s_p}
\end{align}
The width of the side channel is expressed as 
\begin{align}
    w_s=r_s+r_s\cos{\frac{\theta_s^\prime}{2}}=\frac{H_s}{2}\left(1+\cos{\frac{S_s^\prime}{H_s}}\right),
    \label{eq:w_s}
\end{align}
and the length of the straight segment $L_s=L$ is defined as 
\begin{align}
    L_s=L=2r_s\sin{\frac{\theta_s^\prime}{2}}=H_s\sin{\frac{S_s^\prime}{H_s}}.
    \label{eq:L_s}
\end{align}

For the center channel, the radius and subsuming angle of the top and bottom arcs are expressed in terms of the design specifications as 
\begin{align}
    r_c=\frac{1}{2}H_c
    \label{eq:r_c}
\end{align}
and 
\begin{align}
    \theta_c=\frac{S_c}{r_c}=\frac{2S_c}{H_c},
    \label{eq:theta_c}
\end{align}
respectively. 
The angle that subtends the straight segment is 
\begin{align}
    \theta_m=\pi-\theta_c.
    \label{eq:theta_m}
\end{align}
The width of the center channel is 
\begin{align}
    w_c=2r_c\sin{\frac{\theta_c}{2}}=H_c\sin{\frac{S_c}{H_c}},
    \label{eq:w_c}
\end{align}
and the length of the straight segment $L_c=L$ is defined as:
\begin{align}
    L_c=L=2r_c\sin{\frac{\theta_m}{2}}=H_c\cos{\frac{S_c}{H_c}}.
    \label{eq:L_c}
\end{align}

We can then relate the geometries of each channel to each other given the overall width design specifications and their shared straight segments. 
The overall width is 
\begin{align}
    w=w_c+2w_s=H_c\sin{\frac{S_c}{H_c}}+H_s+H_s\cos{\frac{S_s^\prime}{H_s}},
    \label{eq:w}
\end{align}
and Eq. \ref{eq:L_s} and Eq. \ref{eq:L_c} are equated to provide an expression for the term $\frac{S_s^\prime}{H_s}$:
\begin{align}
    L= & L_c=L_s \nonumber \\
    & H_s\sin{\frac{S_s^\prime}{H_s}}=H_c\cos{\frac{S_c}{H_c}} \nonumber \\
    & \frac{S_s^\prime}{H_s}=\sin^{-1}{\left(\frac{H_c}{H_s}\cos{\frac{S_c}{H_c}}\right)}.
    \label{eq:S_s'/H_s}
\end{align}

Eqs. \ref{eq:r_s}-\ref{eq:S_s'/H_s} is then substituted and rearranged to generate equations for fabrication parameters $S_c$, $L$, and $S_s$ in terms of design specifications $H_c$, $H_s$, and $w$, as previously expressed in Section \ref{sec:geometryModel} as Eqs. \ref{eq:S_c}-\ref{eq:beta}. 
Eq. \ref{eq:S_s'/H_s} is substituted into Eq. \ref{eq:w} and rearranged to express $S_c$ as Eq. \ref{eq:S_c}, which can then be substituted into Eq. \ref{eq:L_c} and rearranged to express $L$ as Eq. \ref{eq:L}.
Finally, Eq. \ref{eq:S_c} is substituted into Eq. \ref{eq:S_s'/H_s} and rearranged to express $S_s$ as Eq. \ref{eq:S_s}.

Eq. \ref{eq:S_s} is piecewise to account for the limited range of the arcsin function ($\left[\frac{-\pi}{2}, \frac{\pi}{2}\right]$) compared to the range of $0\le\theta_s\le2\pi$.
The conditions of $w-H_c\sin{\frac{S_c}{H_c}}\le H_s$ and $w-H_c\sin{\frac{S_c}{H_c}}>H_s$ are defined to denote the conditions where $\theta_s\le\pi$ and $\theta_s>\pi$, respectively, in terms of only the design specifications and fabrication parameters. 
These conditions are derived based on the fact that $w_s\le r_s$ if $\theta_s\le\pi$ and $w_s>r_s$ if $\theta_s>\pi$.

\section*{ACKNOWLEDGMENT}

The authors thank Gavin Ng and Booker Schelhaas for their assistance in running human subject experiments.

\addtolength{\textheight}{-12cm}   





\bibliographystyle{IEEEtran}
\bibliography{AirMattressVineRobotBib}

\end{document}